\documentclass[accepted]{uai2022} 
\usepackage[american]{babel}

\usepackage{natbib} 
\usepackage{subcaption}
\usepackage[inline]{enumitem}
\usepackage[linesnumbered,ruled,vlined]{algorithm2e}
\usepackage{wrapfig}
\usepackage{xcolor}
\usepackage{amssymb}
\usepackage{tabularx}
\usepackage[page]{appendix}

\usepackage{hyperref}

\usepackage{mathtools} 
\usepackage{booktabs} 
\usepackage{tikz} 

\title{Forget-me-not! Contrastive Critics for Mitigating Posterior Collapse}

\author[1]{\href{mailto:<sachit.menon@columbia.edu>?Subject=Your UAI 2022 paper}{Sachit Menon}{}}
\author[1]{David Blei}
\author[1]{Carl Vondrick}
\affil[1]{%
    Computer Science Dept.\\
    Columbia University\\
    New York, New York, USA
}
\begin{document}
\maketitle

\begin{abstract}
Variational autoencoders (VAEs) suffer from posterior collapse, where the powerful neural networks used for modeling and inference optimize the objective without meaningfully using the latent representation. We introduce \emph{inference critics} that detect and incentivize against posterior collapse by requiring correspondence between latent variables and the observations. By connecting the critic’s objective to the literature in self-supervised contrastive representation learning, we show both theoretically and empirically that optimizing inference critics increases the mutual information between observations and latents, mitigating posterior collapse. This approach is straightforward to implement and requires significantly less training time than prior methods, yet obtains competitive results on three established datasets. Overall, the approach lays the foundation to bridge the previously disconnected frameworks of contrastive learning and probabilistic modeling with variational autoencoders, underscoring the benefits both communities may find at their intersection.
\end{abstract}

\section{Introduction}

    Variational autoencoders (VAEs) provide an integrated approach for simultaneously performing representation learning and generative modeling. Unlike other approaches, such as generative adversarial networks (GANs), VAEs marry the two steps of probabilistic machine learning -- inference and modeling -- into one framework. They have seen wide success in a number of applications, such as in vision, language, and drug discovery \citep{kingma_auto-encoding_2014, kingma_introduction_2019}. 
    
    VAEs posit a very general model,  where latent variables $\boldsymbol{z}$ give rise to the data $\textbf{x}$. The model thus defines the joint distribution $p_\theta(\textbf{x}, \textbf{z})$, which factorizes as $p(\textbf{z})p_\theta(\textbf{x}|\textbf{z})$. In this factorization, $p(\textbf{z})$ corresponds to a prior (for example, a spherical Gaussian), while $p_\theta(\textbf{x}|\textbf{z})$ defines an exponential family likelihood (usually a Gaussian) with natural parameter dependent on $\textbf{z}$. Much of the power of VAEs as generative models comes from how we define this dependence. Typically, we use the powerful function approximation afforded by neural networks to parametrize this relationship. 
    
    But in a VAE, the power of neural networks can also be its downfall. With a flexible likelihood, the model can learn to abandon the latents entirely.
    This allows the approximate posterior, which is also powered by a neural net, to exactly match the prior. This conspiracy of the inference network and the model network allows the VAE to achieve high values for its objective despite both networks forgetting their respective inputs. While we may achieve some generative modeling goals, this \emph{posterior collapse} phenomenon fails at the goal of representation learning \citep{bowman_generating_2016}. 
    
    This paper proposes a new approach to mitigate posterior collapse. The key idea is that we can use a 
    \emph{critic} to detect posterior collapse and directly incentivize against it. Consider a set of samples of latent variables and the corresponding observations. If posterior collapse has occurred, corresponding latent/observation are independent. The model is not using the latents, and the approximate posterior just produces independent samples from the prior. 
    On the other hand, if we \textit{are} able to pair up corresponding pairs, they must share some information to allow us to do this, and there is no collapse. With this intuition, we create a critic to accomplish precisely this pairing and integrate it into the VAE objective. The critic constrains the neural network to preserve the mutual information between the latent variables and the observations. The resulting generative model must use the information in the data in its posterior of the latent variables. We call this \textit{forget-me-not regularization}.
    
    \begin{figure*}
        \setlength\fboxsep{1pt} 
        \centering
        \includegraphics[width=0.97\linewidth]{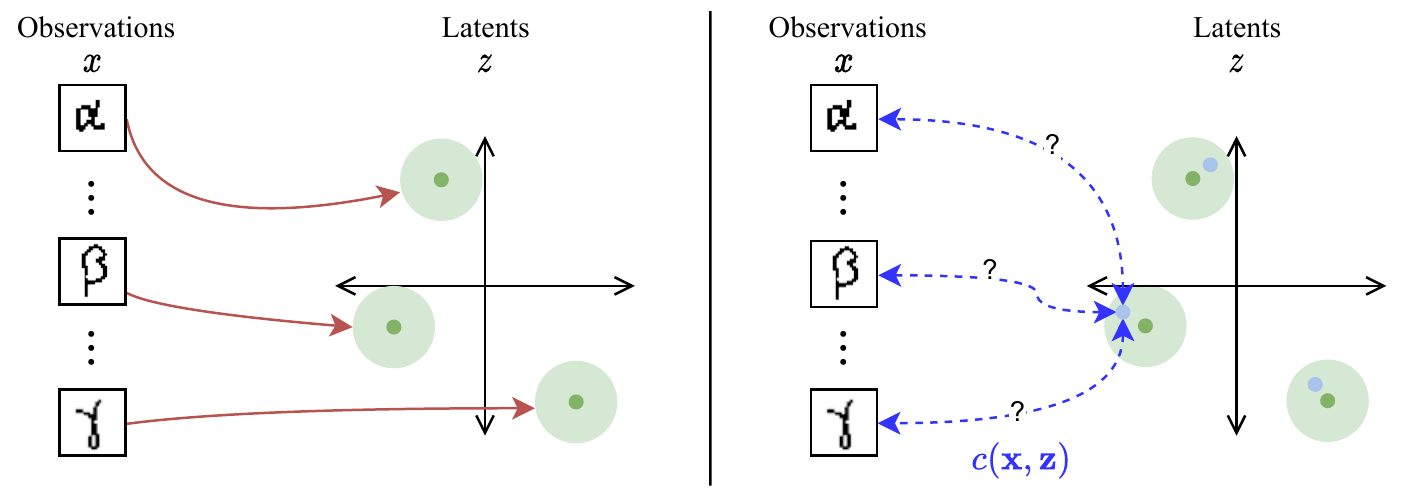}
        \vspace{-1em}
        \caption{An illustration of the critic. On the left, we have the normal \textcolor[RGB]{172,90,83}{variational network} mapping observations to \textcolor[RGB]{108, 135, 85}{variational parameters} (distributions in green). On the right, we show a \textcolor[RGB]{44,65,245}{critic}'s task for a particular \textcolor[RGB]{126, 141, 166}{latent sample} - it must determine which of the blue arrows marks a true pair.}
        \label{fig:illustcritic}
        \vspace{-1em}
    \end{figure*}
  
    
    Inference critics introduce minimal computational overhead and are easy to train. Unlike other posterior collapse strategies (c.f.~ \citep{zhao_infovae_2018}), this critic does not require adversarial training. We are not trying to fool the critic and have it fail its task of distinguishing corresponding pairs (which would actually encourage posterior collapse). Rather, its loss serves as a regularization, biasing the VAE towards solutions where the latent meaningfully relates to its counterpart in observation space. Moreover, this approach  avoids the practical difficulties posed by the `KL annealing’ trick \citep{bowman_generating_2016}, and it does not require multiple experiments to determine a hyperparameter schedule. By connecting the critic to the recent advances in self-supervised contrastive learning \citep{oord_conditional_2016}, we show both theoretically and empirically that the inference critic corresponds to increasing the mutual information between the samples and the latents. 
    
    Experimental results on three standard datasets (across text and image modalities) show that the inference critic provides a robust strategy for mitigating posterior collapse. The approach is also practical, requiring only minimal computational overhead to the standard VAE. It provides significant efficiency gains over established collapse mitigation strategies while achieving competitive performance. 
    
    Our contributions are summarized as: 
    \begin{enumerate*}[label=(\roman*)]
        \item We introduce forget-me-not regularization with inference critics, a self-supervised modification to standard VAEs that substantially reduces effects of posterior collapse.
        \item We show that this modified ELBO formulation directly incentivizes higher mutual information between observations and latents.
        \item We introduce three types of critic: a \textit{neural network critic}, which adds a third neural network to the VAE to act as the critic; a \textit{self critic}, which uses the existing networks to obtain a closed-form optimal solution to the auxiliary task; and a \textit{hybrid critic}, which shares some parameters with the variational network but not all. We contrast these and the effect they have on the final results.
        \item We demonstrate that the method adds less overhead computation time to the standard VAE than other methods for combating posterior collapse.
    \end{enumerate*}

\section{Posterior Collapse in VAEs}
\label{gen_inst}
\vspace{-0.5em}

\subsection{VAE Fundamentals}

    To fit the parameters of a deep generative model, we would ideally maximize the marginal likelihood (the evidence) of the data. However, this is generally an intractable quantity as it involves integrating out the hidden variables.
    Instead, the most common approach is to use variational inference, which allows us to posit a variational family and maximize a tractable lower bound, the ELBO, over its parameters. 
    
    Specifically, the VAE makes use of \textit{amortized} variational inference, which learns a function mapping observations to variational parameters, providing us an approximate posterior over latent variables given observations $q_\phi(\textbf{x}, \textbf{z})$. This function, usually parametrized by a neural network with parameters $\phi$, is shared across data points, hence the amortization. This mechanism for amortized inference is also called the `encoder' in analogy to deterministic autoencoders, with the model referred to as the `decoder'.
    
    There are many equivalent ways to write the ELBO \citep{hoffman_elbo_2016}. Here, we will focus on a couple that illustrate the problem we are addressing and motivate the approach we propose. First consider:
    \begin{equation}
        \begin{aligned}\label{eqn:elbo1}
        \mathrm{ELBO} =&\mathbb{E}_{p_{\mathcal{D}}(\boldsymbol{x})} \mathbb{E}_{q_{\phi}(\boldsymbol{z} \mid \boldsymbol{x})}\left[\log p_{\theta}(\boldsymbol{x} \mid \boldsymbol{z})\right]]]\\ &-\mathrm{KL}\left(q_{\phi}(\boldsymbol{z} \mid \boldsymbol{x}) \| p(\boldsymbol{z})\right)
        \end{aligned}    
    \end{equation}
    where $p_{\mathcal{D}}(\boldsymbol{x})$ is the empirical distribution of observations from the  dataset $\mathcal{D}$. The first term can be thought of as the model conditional likelihood (reconstruction), while the second is the KL divergence between the approximate posterior and the prior.
    
\subsection{Pitfalls in VAE Training}
    
    The form of the ELBO in Equation \ref{eqn:elbo1} illustrates one reason behind the phenomenon of posterior collapse. If the chosen parametrization of the likelihood is flexible enough to learn to always output (a good approximation of) the data distribution, there is no incentive to take a penalty for the second term: the model can keep the first term high even without letting the approximate posterior deviate from the prior. This is one reason behind the phenomenon of posterior collapse: the model does not need the latent code to maximize the likelihood and thus ignores it.
    
   We can provide another expression for the ELBO that provides insight for this case. Consider the variational joint distribution
   \begin{align}\label{eqn:varjoint}
        q_{\phi}(\textbf{x}, \boldsymbol{z}) &= p_{\mathcal{D}}(\boldsymbol{x}) q_{\phi}(\boldsymbol{z} \mid \boldsymbol{x})    
   \end{align}
   
   and aggregate posterior
    \begin{align}\label{eqn:aggpost}
        q_{\phi}(\boldsymbol{z}) &= \mathbb{E}_{p_{\mathcal{D}}(\boldsymbol{x})} q_{\phi}(\boldsymbol{z} \mid \boldsymbol{x})
   \end{align}
   where $p_{\mathcal{D}}(\boldsymbol{x})$ is the data distribution.
   \citep{zhao_infovae_2018,hoffman_elbo_2016,dieng_avoiding_2019,tomczak_vae_2017}.
    Hoffman and Johnson perform `ELBO surgery' \citep{hoffman_elbo_2016} to rearrange the ELBO from Equation \ref{eqn:elbo1} into the following:
    \begin{equation}\label{eqn:elbo2}
            \begin{aligned}
        \mathrm{ELBO} =&\mathbb{E}_{p_{\mathcal{D}}(\boldsymbol{x})} \mathbb{E}_{q_{\phi}(\boldsymbol{z} \mid \boldsymbol{x})}\left[\log p_{\theta}(\boldsymbol{x} \mid \boldsymbol{z})\right]\\ &-\mathcal{I}_{q}(\boldsymbol{x}, \boldsymbol{z})-\mathrm{KL}\left(q_{\phi}(\boldsymbol{z}) \| p(\boldsymbol{z})\right)
        \end{aligned}
    \end{equation}
    where $\mathcal{I}_{q}$ is the mutual information over the variational family. (For a discussion of the variational joint MI $I_q(x;z)$ and the model joint MI $I_p(x;z)$, see Appendix \ref{appendix:mi}.)
    
    In other words, the KL divergence between the posterior and the prior decomposes into a mutual information (MI) penalty and a KL term that encourages matching the aggregate posterior and the prior. Maximizing the ELBO thus explicitly discourages high mutual information between observations and latents, pushing them towards independence. As \citep{hoffman_elbo_2016} show, decreasing the value of the MI does not impact the likelihood term, which tends to dominate. As such, the objective enables the flexible neural nets to achieve solutions exhibiting posterior collapse.
    
    This loss of information means that the latents will not be informative about the observations (and thus cannot be useful representations). In the next section, we introduce a method to explicitly prevent this information loss or `forgetting'. By incorporating a regularization term that incentivizes higher MI into the ELBO, we will counteract the effect of the MI penalty from Equation \ref{eqn:elbo2}.
    
  
\section{Forget-me-not Regularization} \label{section:method}
    
    We will employ a critic that imposes a penalty on the objective if observations and their corresponding latents cannot be distinguished from non-corresponding pairs. The intuition is that this matching is only possible if there is information shared between observations and latents. 
    
    \subsection{The Inference Critic}
    Consider a batch of samples from the empirical data distribution $x_0, \ldots, x_k$, and a corresponding batch of latent samples $z_0, \ldots, z_k$ (by encoding the $x_i$). Every pair with the correct correspondence comes from the  variational joint distribution $q(\textbf{x}, \textbf{z})$, while non-corresponding pairs are independent and come from the product of marginals ($\textbf{z}$ via ancestral sampling) \citep{alemi_fixing_2018}. 
    Formally:
    \begin{equation}\label{eqn:sampling}
       (z_i, x_j) \sim 
        \begin{cases} 
          q_\phi(\textbf{z}, \textbf{x}) & i = j \\
          q_\phi(\textbf{z})p_\mathcal{D}(\textbf{x}) & i \neq j
       \end{cases}
    \end{equation}

    If we can distinguish the joint distribution from the product of marginals, there must be some dependence between $\textbf{x}$ and $\textbf{z}$. Given samples from both distributions, the critic will try to pick out which pairs belong to which distribution. The more successful it is, the more different the distributions must be, and therefore the more $\textbf{x}$ and $\textbf{z}$ must be related. 
    
    
    The classifier, which we will denote $f$, needs to distinguish between the joint and the product of marginals by directly \textit{contrasting} the correct pairings from the incorrect ones.
    We know that for every observation $x_i$, there is one latent in the batch $z_i$ that corresponds to it (and vice versa). We  also know that the others \textit{should not}. 
    This corresponds to softmax classification \citep{bishop_pattern_2006}, or a probabilistic classifier with a categorical likelihood. 
    $f$ maximizes the objective: 
    \begin{equation}\label{eqn:classification}
        c(\textbf{x},\textbf{z}) = \mathbb{E} \left[\log \frac{f\left(x^{+}, z^{+}\right)}{\sum_{x \in X } f\left(x, z^{+}\right)}\right]
    \end{equation}
    which is the critic's expected value for the true corresponding pairs (denoted by the $+$) relative to all the other (non-corresponding) pairs, across pairs.\footnote{Note writing $f(x,z)$ as $\exp(f_0(x,z))$ recovers the softmax classifier exactly.} (The notation within the sum is in reference to a particular positive $z^{+}$; we can think of it as considering a particular $z^{+}$ and trying to find the associated $x^{+}$ among the options for $x$, then taking the expectation over $z$. This is symmetric with respect to choosing an $x^{+}$ and finding the associated $z^{+}$.) 
    
    \begin{algorithm}[h!]
      \KwIn{Dataset $\mathcal{D}$, batch size $K$, initial VAE parameters $\theta$, initial critic $f$ parametrized by $\psi$, regularization weight $\lambda$}
      \While{not converged}
       {
       		Sample from $p_\mathcal{D}(\textbf{x})$ $K$ times to obtain $(x^{(i)})_{i=1}^K$
       		
       		Sample $z^{(i)}\sim q_\phi(\textbf{z}|x^{(i)}\ \forall i\in \{1,\ldots,K\}$
       		
       		Compute $\mathcal{L}_0 = \sum_{i=1}^K \text{ELBO}(\theta, \phi, x^{(i)}, z^{(i)})$ per Eqn \ref{eqn:elbo1} \tcp*{standard minibatch ELBO}
       		
       		Compute $f_\psi (x_i,z_j) \forall i,j$ \tcp*{inference critic values}
       		
       		$\mathcal{L}_1 \leftarrow c(\textbf{x},\textbf{z})$ per Eqn \ref{eqn:classification} \tcp*{inference critic objective}
       		
       		$\mathcal{L} \leftarrow \mathcal{L}_0 + \lambda*\mathcal{L}_1$
       		
       		Perform gradient update for VAE parameters 
       		
       		Perform gradient update for critic parameters 
       }
    \caption{Forget-me-not regularization with neural network inference critic.}\label{alg:nncrit}
    \end{algorithm}
    
    Crucially, this objective constitutes a lower bound of the mutual information, shown by \citet{oord_representation_2019} in the context of self-supervised representation learning. By maximizing Equation \ref{eqn:classification}, the classifier approximates the density ratio between the joint distribution and product of marginals \citep{oord_representation_2019,song_multi-label_2020}, which is precisely the ratio appearing in the mutual information.
    Therefore, by optimizing the parameters of the VAE with this objective added to the ELBO as a regularizer, we push up on this lower bound. This push increases the MI between the latents and the observations, effectively mitigating posterior collapse as desired.
    
    This critic establishes a tight connection to the contrastive learning literature, in particular the InfoNCE loss from Contrastive Predictive Coding (CPC) \citep{oord_representation_2019}, enabling the many advances in self-supervised learning to be applied to VAEs. See Appendix \ref{appendix:multimi} for details of how this connection and bound apply. Note that, in the MI estimation literature, classifiers that estimate the mutual information like CPC are sometimes also referred to as critics \citep{poole_variational_2019}. We intentionally overload this word here: the inference critics of this paper critique the variational inference optimization.

    \subsection{Regularization}

    In order to prevent posterior collapse, we want to maximize the mutual information between the latents and the samples. We integrate the penalty of the critic, which aims to maximize $c(\textbf{x}, \textbf{z})$, to the ELBO:
    \begin{equation}\label{eqn:elbo-ours}
        \begin{aligned}
        \mathrm{ELBO_{CRITIC}} &=\mathbb{E}_{p_{\mathcal{D}}(\boldsymbol{x})} \mathbb{E}_{q_{\phi}(\boldsymbol{z} \mid \boldsymbol{x})}\left[\log p_{\theta}(\boldsymbol{x} \mid \boldsymbol{z})\right] \\&-\mathrm{KL}\left(q_{\phi}(\boldsymbol{z} \mid \boldsymbol{x}) \| p(\boldsymbol{z})\right) + c(\textbf{x}, \textbf{z})
        \end{aligned}    
    \end{equation}
    We optimize the parameters for the variational family, model, and critic jointly; the critic prevents the usual conspiracy between the model and variational family over the course of training, avoiding collapse.     Algorithm \ref{alg:nncrit} illustrates the training procedure.
    
    \begin{figure}
        \centering
        \includegraphics[width=0.9\linewidth]{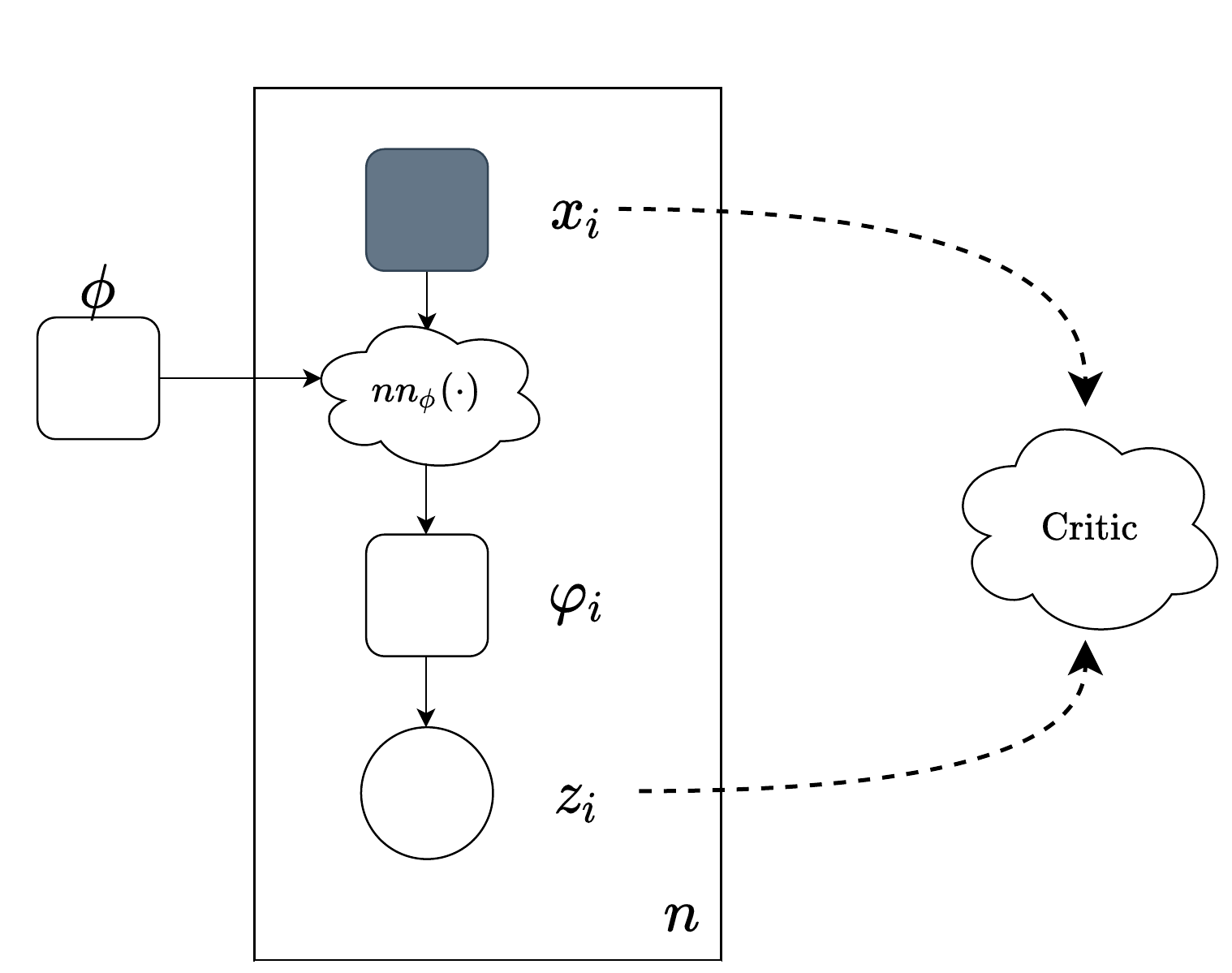}
        \caption{Forget-me-not regularization in reference to the graphical models acted on. Dashed lines show computation flow. The inference critic works against information loss in the variational family, trying to ensure it can tell $z_i$ and $x_i$ correspond. It does this by solving a classification task: among a batch of different repetitions of these variables, classify which belong together. This task can only be solved if there is shared information between variables belonging to the same subgraph (repeated by the plate) to distinguish them from those belonging to others. } 
        \label{fig:critics}
    \end{figure}
    
    Notice that the mutual information appears in both the KL-divergence term and the critic term. When the true mutual information is equal to our estimate, this corresponds to matching the marginals for $z$ and the mutual information term from Equation \ref{eqn:elbo2} is cancelled out by the regularization. This approach does not need to resort to adversarial training techniques that are difficult to use in practice. Instead, the critic provides a straightforward mechanism to mitigate posterior collapse by causing an increase in the mutual information term. 
    
    We apply this regularization across the variational family - an \textit{inference} critic, illustrated in Figure \ref{fig:critics}. 
    See Appendix \ref{appendix:critics} for discussion of its potential application across the model. 
    Furthermore, while here we consider the problem of posterior collapse by examining the ELBO for the VAE, posterior collapse is a more general phenomenon, observed in many different models when using amortized variational inference with deep neural networks. In theory, this approach would work on any such problem with amortized inference. 

    \subsection{Types of Inference Critics}
    
    This framework affords multiple types of critics (as forms for $f$ in Equation \ref{eqn:classification}) that correspond to the mutual information between the latents and the samples. We propose three types of inference critics. 
    
    The most straightforward critic is a \textbf{neural network critic}. This critic uses a third network, entirely separate from those used in the VAE, to implement the critic. The choice of the network design depends on the structure of the modality. For example, we could use an embedding layer followed by an LSTM for text data. 
    
    We also propose a \textbf{self-critic} that uses the variational family as its own critic, providing the tighest estimate of the mutual information. This originates from an idea in noise contrastive estimation: if we have a tractable conditional (as is the case with the variational family in the VAE), we can directly use it to estimate the mutual information, and in particular the tightest estimate of the MI will be found by using $f(x,z) = \log q(z|x)$  \citep{poole_variational_2019}. This formulation incentivizes the log-likelihood under the conditional to be highest for samples that actually belong together, as we would desire. This critic also has the advantage of not requiring additional parameters.
    
    The \textbf{hybrid critic}, as the name suggests, is in-between the self-critic and the neural network critic. Rather than use an entirely separate neural network, this critic shares early layers with the variational network, after which point it has its own parameters. This approach can be well-suited to text data, where we may wish to share the embedding weights between the variational network and critic but have them separate past that. This presents a way to compromise between using no additional parameters (as in the self-critic) and using a whole additional network's worth of parameters (as in the neural network critic).

 \begin{table*}[h!]
        \centering
        \centerline{
        \scalebox{0.9}{
        \begin{tabular}{lcccc|ccccc}
        \hline & & \multicolumn{2}{c} { Yahoo } & & & \multicolumn{2}{c} { Yelp } & & \\
        Model & NLL & KL & MI ($I_q$) & AU & NLL & KL & MI ($I_q$) & AU \\
        \hline VAE & 
        328.9 (0.1) & 0.0 (0.0) & 0.0 (0.0) & 0.0 (0.0) &
        358.3 (0.2) & 0.0 (0.0) & 0.0 (0.0) & 0.0 (0.0)\\
        SA-VAE \citep{kim_semi-amortized_2018} & 
        329.2 (0.2) & 0.1 (0.0) & 0.1 (0.0) & 0.8 (0.4) &
        357.8 (0.2) & 0.3 (0.1) & 0.3 (0.0) & 1.0 (0.0) \\
        Skip-VAE \citep{dieng_avoiding_2019} & 
        328.7 (0.3)  & 0.22 (0.1) & 0.0 (0.0) & 7.0 (0.6) &
        358.1 (0.3) & 0.15 (0.0) & 0.0 (0.0) & 4.6 (0.5) \\
        Lagging-VAE \citep{he_lagging_2019} & 
        \textbf{328.2} (0.2) & 5.6 (0.2) & 3.0 (0.0) & 8.0 (0.0) &
        \textbf{356.9} (0.2) & 3.4 (0.3) & 2.4 (0.1) & 7.4 (1.3) &\\
        VAE + Inference Critic (Self) & 
        328.7 (0.2) & 3.6 (0.1) & 2.6 (0.0) & 3.0 (0.0) &
        358.2 (0.2) & 3.8 (0.1) & 2.7 (0.0) & 3.0 (0.0) &\\
        VAE + Inference Critic (Hybrid) & 
        \textbf{328.2} (0.1) & 4.3 (0.1) & 2.8 (0.0) & \textbf{11.0} (0.4)&
        357.7 (0.2) & 4.0 (0.1) & 2.8 (0.0) & 7.0 (0.0) &\\
        VAE + Inference Critic (NN) & 
        338.9 (0.4) & \textbf{17.5} (1.1) & \textbf{3.3} (0.0) & 8.0 (1.0) & 
        370.5 (0.5) & \textbf{18.6} (1.8) & \textbf{3.2} (0.1) & \textbf{12.0} (2.0)& 
        \end{tabular}
        }
        }
        \caption{Quantitative results on the Yahoo and Yelp  text corpora.   Each critic improves on collapse metrics when added to the standard VAE with no other changes. Results for comparison without KL annealing were referenced from \citet{he_lagging_2019} or re-implemented in the same framework and are averages of 5 runs, with standard deviation given in parentheses. (We follow the training details in the aforementioned methods, running until convergence is achieved on the validation ELBO.)}
        \label{table:textresults}
        \centering
        \begin{tabular}{lcccc}
        \hline & & \multicolumn{2}{c} { Omniglot } &  \\
        Model & NLL & KL & MI ($I_q$) & AU \\
        \hline VAE & 
        89.41 (0.04) & 1.51 (0.05) & 1.43 (0.07) & 3.0 (0.0)\\
        SA-VAE \citep{kim_semi-amortized_2018} & 
        89.29 (0.04) & 2.55 (0.05) & 2.20 (0.03) & 4.0 (0.0)\\
        Skip-VAE \citep{dieng_avoiding_2019} & 
        89.41 (0.05) & 1.75 (0.20) & 1.61 (0.10) & 3.0 (0.4)\\
        Lagging-VAE \citep{he_lagging_2019} & 
        \textbf{89.05} (0.05) & 2.51 (0.14) & 2.19 (0.08) & 5.6 (0.5) \\
        VAE + Inference Critic (Self) & 
        89.18 (0.04) & 6.30 (0.12) & 3.75 (0.04) & 11.0 (1.0)\\
        VAE + Inference Critic (Hybrid) & 
        89.16 (0.04) & 6.41 (0.15) & 3.78 (0.03) & 13.0 (0.7) \\
        VAE + Inference Critic (NN) & 
        89.24 (0.05) & \textbf{7.66} (0.14) & \textbf{3.82} (0.04) & \textbf{28.0} (0.0)
        \end{tabular}
        \caption{Quantitative results on the Omniglot image dataset. We find each critic improves on collapse metrics when added to the standard VAE with no other changes. Results for comparison without KL annealing were referenced from \citep{he_lagging_2019} or re-implemented in the same framework and are averages of 5 runs, with standard deviation given in parentheses. (We follow the training details in the aforementioned methods, running until convergence is achieved on the validation ELBO.)}
        \label{table:imgresults}
    \end{table*}

\section{Experiments}

The basic objective of our experiments is to analyze inference critics under posterior collapse. We report results across three established image and text datasets.

\subsection{Common Experimental Setup}

We follow a common setup throughout our experiments. Since the method is compatible with the VAE family, it can also be added on top of existing methods to mitigate posterior collapse. We conducted experiments adding forget-me-not regularization to a standard VAE to assess whether the theoretical improvements yielded empirical benefits, following \citep{he_lagging_2019} and collapse metrics from \citep{dieng_avoiding_2019}. We measure measuring the approximate negative log-likelihood (NLL, via $500$ importance samples - this gives a tighter bound for the evaluation than the ELBO), the ELBO's KL term, a Monte Carlo estimate of the MI under the variational joint $I_q(x;z)$, and the number of active units (AU) \citep{burda_importance_2016} on held-out data.

We report results for the Yahoo, Yelp, and Omniglot datasets, allowing systematic comparison to prior work \citep{he_lagging_2019,kim_semi-amortized_2018,dieng_avoiding_2019}. As there are both visual and text datasets, we use existing, appropriate  neural network architectures for each modality, which we describe next to each experiment. For all datasets, we follow the standard train/validation splits provided by the original dataset authors. We evaluated all results on a single NVIDIA GeForce RTX 2048 GPU. 

\textbf{Baselines:} For all experiments, we compare against multiple established baselines. We use the standard \textbf{VAE \citep{kingma_auto-encoding_2014}} without any additional strategy for handling posterior collapse. We also compare to \textbf{SA-VAE \citep{kim_semi-amortized_2018}}. Rather than solely relying on amortized inference to obtain variational parameters, the semi-amortized VAE (SA-VAE) uses amortized inference to obtain an initialization and updates the parameters directly from that point. Finally, we compare to the \textbf{Lagging-VAE \citep{he_lagging_2019}}, which aggressively updates the variational family many more times (e.g. 50x) as frequently as the model. We choose this as representative as Lagging-VAE holds the previous state-of-the-art on both modeling and posterior collapse metrics without requiring use of the KL annealing trick, making it a competitive baseline. 

\subsection{Evaluation Metrics}

We evaluate all models and baselines using the standard metrics for evaluating the posterior collapse of VAEs. We report the following:

\textbf{Negative Log Likelihood (NLL):} The negative log-likelihood indicates the modeling performance on held-out 
data. A smaller NLL indicates that the model generalizes to data samples well.

\textbf{KL-Divergence (KL):} The KL-divergence term of the ELBO in Equation \ref{eqn:elbo1} is a commmon indicator of collapse. If we obtain a good ELBO but the KL term is low, the optimization progress comes from the model likelihood (the first term of Equation \ref{eqn:elbo1}). In this case, especially if the KL is at or near zero, the posterior matches the prior too well (suggesting collapse has occurred). 

\textbf{Mutual Information (MI):} 
The estimate of the mutual information across the variational family, $I_q(x;z)$, aims to estimate if the latents have become independent of the data (over the variational joint). We compute this estimate as the difference of the previously-described KL term and the marginal KL term from equation \ref{eqn:elbo2} per \citet{hoffman_elbo_2016}. Both KL terms are obtained by Monte Carlo. The first is obtained naturally in computation of the ELBO, while the second can be computed via ancestral sampling (sampling from the dataset, then the approximate posterior) again as in \citep{hoffman_elbo_2016}. However, as pointed out by \citep{he_lagging_2019}, this estimate is biased - specifically, it is an upper bound. 

\textbf{Active Units (AU):} A standard metric from prior work, the number of active units provides a measure of how many latent dimensions are active, which is specifically how many of the stochastic units show any variation when the input varies. If few are active, we are likely collapsed. Activity is measured by $A_{z}=\operatorname{Cov}_{\mathbf{x}}\left(\mathbb{E}_{z \sim q(z \mid \mathbf{x})}[z]\right)$, with a unit considered active if its activity is above some threshold (we follow \citep{dieng_avoiding_2019} and \citep{he_lagging_2019} with a threshold of $0.01$). 

    \begin{table}[t!]
        \centering
        \begin{tabularx}{\linewidth}{lcc}
        \hline & \multicolumn{2}{c} {Running Time}  \\
        Model & Factor & Absolute (Hrs) \\
        \hline VAE & 
        1.00 & 3.5 \\
        Lagging-VAE & 
        3.6 & 12.7 \\
        VAE + Inf. Critic (Hybrid) & 
        1.06 & 3.7
        \end{tabularx}
        \caption{Speed comparison results. We report wall clock time (`Absolute') and factor increase over the baseline (`Factor') on the Yahoo benchmark. Adding an inference critic adds minimal overhead. Per the experimental procedure used in \citep{he_lagging_2019}, we run to convergence on the validation ELBO; the standard VAE converges after 54 epochs, Lagging-VAE converges after 49 epochs, and the VAE with a hybrid critic converges after 54 epochs. For the same number of epochs, the speed difference only increases.}
        \label{table:timeresults}
        \vspace{-1.5em}
    \end{table}

\subsection{Results on Text}

\textbf{Experimental Setup:} In this experiment, we evaluate on the Yahoo and Yelp benchmarks \citep{yang_improved_2017}. All methods use the standard train/val/test splits and follow the experimental protocol in \citep{he_lagging_2019}, fully described in Appendix \ref{appendix:protocol}. 
For the neural network architecture, this is a 1-layer LSTM with learnable embeddings for the variational network and for the model network. Following this protocol, all methods use a 32-dimensional latent space and a batch size of 32. 

\textbf{Quantitative Results:} The quantitative results in Table \ref{table:textresults} show that all critics substantially improve on the collapse metrics compared to the baseline on both datasets, showing forget-me-not regularization is able to significantly mitigate posterior collapse on text data. 

Our results show that different critics have different behaviors. As the self-critic optimally solves the auxiliary task at each step, if there is any information shared between observations and latents, it is able to pair them up successfully. On the other hand, the neural network critic is entirely separate from the variational network and is decidedly sub-optimal at the auxiliary task - and thus needs more information to be shared, as that increases its chances of finding a way to tie together corresponding pairs. As the experimental results show, this encourages less collapse. In particular, the neural network critic reaches close to the theoretical maximum $I_q$ increase offered by applying an inference critic with the batch size used ($\log(32) \approx 3.4$, see Appendix \ref{appendix:multimi}). At the same time, this may lead to solutions that do not perform as well along the NLL, because there is a much stronger pull to not collapse. The hybrid critic reaches a balance between these: by not being optimal at solving the auxiliary task, it pulls more strongly away from collapse, but by sharing some parameters with the variational network, it is able to reach better modeling solutions than the entirely separate neural network critic, actually improving the NLL slightly over the standard VAE. It is interesting that the neural network critic, which uses an entirely separate set of parameters and thus should be strictly more flexible, fails to reach such an optima; we hypothesize this comes from difficulty in optimization, which is made easier when the weights are tied. None of the critics require additional hyperparameter scheduling, such as to control KL annealing.

\begin{figure}
    \centering
    \includegraphics[width=0.9\linewidth]{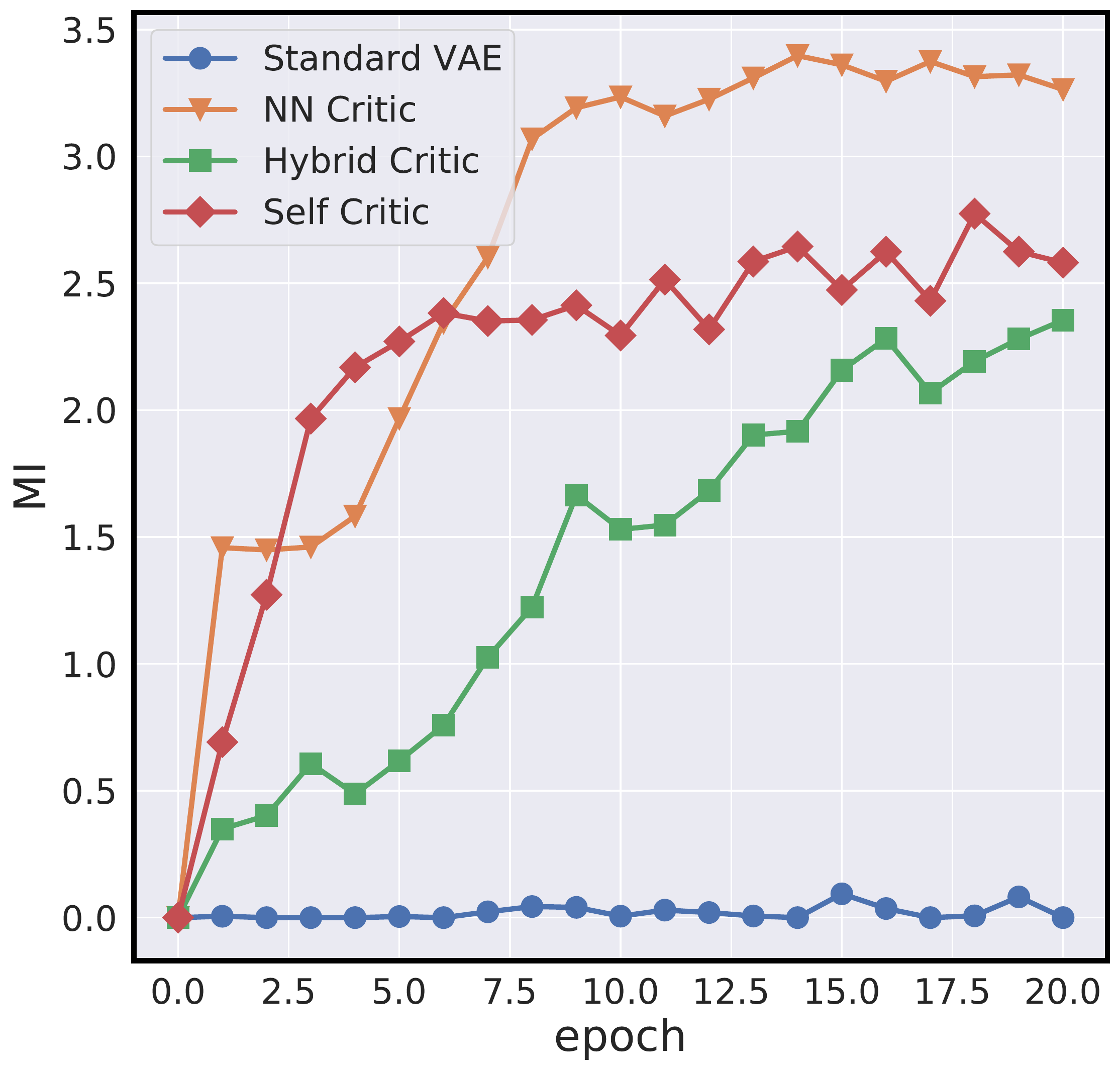}
    \caption{Comparison of mutual information across the variational family ($I_q$) for various critics vs baseline over the first 20 epochs of training on the Yahoo benchmark. This is cropped for clarity; the plot over the entirety of training can be seen in Appendix \ref{appendix:miexpanded}.}
    \vspace{-2em}
    \label{fig:mi_comparison}
\end{figure}

Figure \ref{fig:mi_comparison} shows that all critics progressively push the mutual information up over the course of training. In contrast, the standard VAE remains collapsed throughout training. 

\subsection{Results on Images}

\textbf{Experimental Setup:} We next evaluate on the 
Omniglot benchmark \citep{lake_human-level_2015} with the provided train/val/test split. All methods follow the experimental protocol in \cite{he_lagging_2019}, fully described in the Appendix.  For the architecture, all methods use a ResNet \citep{he_deep_2015} for the variational network and a 13-layer Gated PixelCNN \citep{oord_conditional_2016} for the model.
All methods use a 32-dimensional latent space with a batch size of 50. 

\textbf{Quantitative Results:} The results on images in Table \ref{table:imgresults} show a similar trend to the results on text, suggesting that the approach is robust across modalities. We find inference critic is able to mitigate collapse for images as well, improving substantially on collapse metrics compared to the baseline. 

Furthermore, the relative quantitative behavior of different critics remains the same on the images as well as text. 
In particular, the neural network critic reaches close to the theoretical maximum $I_q$ increase offered by applying an inference critic with the batch size used ($\log(50) \approx 3.9$. See Appendix \ref{appendix:multimi}) for more details. Finally, unlike for text, all critics are able to improve NLL over the baseline. It is interesting that the regularization actually seems to help the original objective; this behavior is unlike that of approaches like Beta-VAE \citep{higgins_beta-vae_2016}. We hypothesize that while the collapsed solution may be `easy' to arrive to with SGD, there exist better optima that actually use the latents, which inference critics drive the parameters towards. 
\subsection{Efficiency and Running Time Performance}

A key advantage of this approach is that it adds minimal overhead to standard VAE training. As Table \ref{table:timeresults} shows, the hybrid critic completes training in only 1.06x the time the standard VAE takes - 3.7 hours vs 3.5 hours in wall-clock time on one NVIDIA RTX 2048. This comes at no performance trade-off: the hybrid critic improves along the collapse metrics KL, MI, and AU over the standard baseline with no reduction in NLL. The time performance is particularly important as the overhead of previous approaches to avoiding posterior collapse is typically too high for large-scale problems. While Lagging-VAE does help against posterior collapse, it took 12.7 hours to train (3.4x the time of the hybrid critic). Other approaches are reported in the literature to require further computation - for example, SA-VAE takes between 4 and 8 times as long as Lagging-VAE in \citep{he_lagging_2019}.

How can our method have such low overhead? Contrastive learning often can be quite expensive. The 6\% overhead is made possible with the hybrid critic, as it shares most of its parameters with the inference network. The bulk of the computation is thus only done once, with the overhead introduced when the two diverge. Additionally, with inspiration from current work in self-supervised learning we employ critics that are separable \citep{poole_variational_2019}, meaning they do not need to take every $(x,z)$ pair (which would be $n^2$ forward passes) but only the x. This leads to quadratic big-O speedup compared to other approaches that need to process every potential pairing.

\section{Related Work}
    The most relevant line of study related to this work are methods that try to avoid posterior collapse - especially those that do so by increasing the mutual information. \citep{bowman_generating_2016} identify the problem of posterior collapse for VAEs endowed with powerful generators. They prescribe `KL annealing', or slowly increasing the KL-penalty in Equation \ref{eqn:elbo1} at the start of training. From Equation \ref{eqn:elbo2}, we can interpret this as slowly ramping up the MI penalty inherent to the ELBO. Approaches like \citep{chen_variational_2017}, \citep{gulrajani_pixelvae_2016}, \citep{yang_improved_2017} modify the architecture of the generative model to reduce its flexibility, with the hope that this will prevent it from finding solutions that `forget'. \citep{dieng_avoiding_2019} instead adds skip connections to the VAE model network to increase $I_p(x;z)$. On the other hand, \cite{wang_posterior_2021} connect posterior collapse to latent variable non-identifiability. This also provides another perspective on other methods, suggesting methods that modify the objective (such as this work) may prevent collapse by encouraging solutions with identifiable latent variables.
    
    Other approaches aim to explicitly encourage higher mutual information by modifying the objective, towards the goal of `fixing a broken ELBO' as identified by \cite{alemi_fixing_2018}. Forget-me-not regularization falls into this category. (Note that we do not claim novelty in the general idea of modifying the objective to avoid MI loss; rather, forget-me-not regularization presents a new way of accomplishing this with distinct advantages in optimization ease, simplicity, and speed.) \cite{zhao_infovae_2018} introduce an explicit marginal-KL penalty which can be traded off with the usual KL term in Equation \ref{eqn:elbo1}. They do this either with an adversarial classifier that tries to guess if a sample is from the aggregate posterior or the prior, Stein variational gradient descent \citep{liu_stein_2019}, or kernel-based methods (e.g., MMD \citep{gretton_kernel_2012}), each of which is difficult to work with. They show adversarial autoencoders \citep{makhzani_adversarial_2016} are a special case of the former. \cite{razavi_preventing_2019} enforce a minimum KL instead of using a penalty.
    
    Wasserstein autoencoders \citep{tolstikhin_wasserstein_2019} provide another approach to marginal matching, aiming to avoid the traditional KL term (and its associated MI penalty) entirely by matching marginals with the Wasserstein distance (but this also requires adversarial training or MMD - it is also a generalization of adversarial autoencoders). \cite{phuong_mutual_2018} adds an explicit penalty for the MI with the Barber and Agakov lower bound \citep{barber_im_2004}, but this is not fully differentiable and requires use of the high-variance REINFORCE gradient estimator.  Closely related to this work is \cite{rezaabad_learning_2020}, which also uses an auxiliary network on batches of samples (from the variational joint) to obtain a penalty term; instead of solving a straightforward classification problem as done here, they use these to try to estimate the dual form of the mutual information. VAE-MINE \citep{qian_enhancing_2019} use the MINE bound (a cousin of the CPC bound) to aim to increase MI, but as pointed out by \cite{poole_variational_2019}, the way it is computed does not constitute a correct MI bound. (See Appendix \ref{appendix:vaemine} for more detailed discussion of VAE-MINE.)   Along different lines, \cite{kim_semi-amortized_2018} address posterior collapse by using the inference network's outputs as an initialization for SVI, creating a hybrid procedure between amortized and non-amortized inference. \cite{he_lagging_2019} modify the optimization procedure to take more steps for the inference network than the model network.
    
    The idea of including an auxiliary classification task to generative models has a long history. The most famous of these are GANs \citep{goodfellow_generative_2014}, which train an auxiliary classifier adversarially on the empirical data distribution and the model data distribution, aiming to make these indistinguishable. \cite{uehara_generative_2016} interprets this in the framework of density ratio estimation. Bayesian GANs \cite{tran_hierarchical_2017} modify these with Bayesian neural networks for Bayesian inference. Hybrids between VAEs and GANs have been introduced in many forms \citep{larsen_autoencoding_2016,donahue_adversarial_2017,srivastava_veegan_2017,mescheder_adversarial_2017}.  All of these use the auxiliary classifier adversarially, to encourage distribution matching, whereas we are encouraging distributions \textit{not} to match - specifically, we are encouraging dependence between latents and observations by making the joint different from the product of marginals. \cite{aneja_contrastive_2021} use an auxiliary classifier for noise-contrastive estimation between the prior and the aggregate posterior to address the `prior hole' problem. Rather than auxiliary classifiers, \cite{seybold_dueling_2019} add auxiliary decoders aiming to avoid collapsed optima.
    
    This work provides a foundation for connecting advances in contrastive learning with the VAE framework, giving us control over \textit{what} information is preserved (vs what the representations are invariant to).
    CPC \citep{oord_representation_2019} introduces the InfoNCE mutual information bound, applying it to representations obtained by applying autoregressive models to sequential data. \citep{wu_unsupervised_2018} proposes the instance discrimination problem, which suggests matching observations with encodings of perturbed versions of the same observation - for example, two different augmentations of an image, like brightness shifts - learning representations that are invariant to these factors. MoCo \citep{he_momentum_2020} proposes a `momentum queue' to hold encodings from recent batches to increase the number of `negatives' being compared to substantially; it surpasses the performance of representations learned by supervised neural networks on various computer vision tasks. 
    \cite{zhu_eqco_2020} allows for contrastive learning without large numbers of negatives via a small modification to the InfoNCE objective, that provides a more practical bound on the mutual information. \cite{tschannen_mutual_2020} provides connections of these techniques and associated mutual information bounds to metric learning, which could also be an interesting perspective to bring to representation learning with VAEs. \vspace{-1em}
\section{Conclusions}
We present a new method for protecting against posterior collapse in VAEs. In doing so, we establish a connection between VAEs and contrastive representation learning. We show inference critics increase the mutual information between latents and observations by maximizing the CPC lower bound. Experiments on three datasets show the effectiveness of the approach, with significant efficiency gains. 


\bibliographystyle{plainnat}
\bibliography{references.bib}

\begin{thebibliography}{48}
\providecommand{\natexlab}[1]{#1}
\providecommand{\url}[1]{\texttt{#1}}
\expandafter\ifx\csname urlstyle\endcsname\relax
  \providecommand{\doi}[1]{doi: #1}\else
  \providecommand{\doi}{doi: \begingroup \urlstyle{rm}\Url}\fi

\bibitem[raz(2019)]{razavi_preventing_2019}
Preventing {Posterior} {Collapse} with delta-{VAEs}, January 2019.
\newblock URL \url{http://arxiv.org/abs/1901.03416}.
\newblock Number: arXiv:1901.03416 arXiv:1901.03416 [cs, stat].

\bibitem[sey(2019)]{seybold_dueling_2019}
Dueling {Decoders}: {Regularizing} {Variational} {Autoencoder} {Latent}
  {Spaces}, May 2019.
\newblock URL \url{http://arxiv.org/abs/1905.07478}.
\newblock Number: arXiv:1905.07478 arXiv:1905.07478 [cs, stat].

\bibitem[ane(2021)]{aneja_contrastive_2021}
A {Contrastive} {Learning} {Approach} for {Training} {Variational}
  {Autoencoder} {Priors}, November 2021.
\newblock URL \url{http://arxiv.org/abs/2010.02917}.
\newblock Number: arXiv:2010.02917 arXiv:2010.02917 [cs, stat].

\bibitem[Alemi et~al.(2018)Alemi, Poole, Fischer, Dillon, Saurous, and
  Murphy]{alemi_fixing_2018}
Alexander~A Alemi, Ben Poole, Ian Fischer, Joshua~V Dillon, Rif~A Saurous, and
  Kevin Murphy.
\newblock Fixing a {Broken} {ELBO}.
\newblock page~21, 2018.

\bibitem[Barber and Agakov(2004)]{barber_im_2004}
David Barber and Felix Agakov.
\newblock The {IM} {Algorithm} : {A} variational approach to {Information}
  {Maximization}.
\newblock In \emph{{NeurIPS}}, page~8, 2004.

\bibitem[Bishop(2006)]{bishop_pattern_2006}
Christopher~M. Bishop.
\newblock Pattern recognition and machine learning, 2006.
\newblock URL \url{https://cds.cern.ch/record/998831}.
\newblock ISBN: 9781493938438 9780387310732 Publisher: Springer.

\bibitem[Bowman et~al.(2016)Bowman, Vilnis, Vinyals, Dai, Jozefowicz, and
  Bengio]{bowman_generating_2016}
Samuel~R. Bowman, Luke Vilnis, Oriol Vinyals, Andrew~M. Dai, Rafal Jozefowicz,
  and Samy Bengio.
\newblock Generating {Sentences} from a {Continuous} {Space}.
\newblock \emph{arXiv:1511.06349 [cs]}, May 2016.
\newblock URL \url{http://arxiv.org/abs/1511.06349}.
\newblock arXiv: 1511.06349.

\bibitem[Burda et~al.(2016)Burda, Grosse, and
  Salakhutdinov]{burda_importance_2016}
Yuri Burda, Roger Grosse, and Ruslan Salakhutdinov.
\newblock Importance {Weighted} {Autoencoders}.
\newblock \emph{arXiv:1509.00519 [cs, stat]}, November 2016.
\newblock URL \url{http://arxiv.org/abs/1509.00519}.
\newblock arXiv: 1509.00519.

\bibitem[Chen et~al.(2017)Chen, Kingma, Salimans, Duan, Dhariwal, Schulman,
  Sutskever, and Abbeel]{chen_variational_2017}
Xi~Chen, Diederik~P. Kingma, Tim Salimans, Yan Duan, Prafulla Dhariwal, John
  Schulman, Ilya Sutskever, and Pieter Abbeel.
\newblock Variational {Lossy} {Autoencoder}.
\newblock \emph{arXiv:1611.02731 [cs, stat]}, March 2017.
\newblock URL \url{http://arxiv.org/abs/1611.02731}.
\newblock arXiv: 1611.02731.

\bibitem[Dieng et~al.(2019)Dieng, Kim, Rush, and Blei]{dieng_avoiding_2019}
Adji~B. Dieng, Yoon Kim, Alexander~M. Rush, and David~M. Blei.
\newblock Avoiding {Latent} {Variable} {Collapse} {With} {Generative} {Skip}
  {Models}.
\newblock \emph{arXiv:1807.04863 [cs, stat]}, January 2019.
\newblock URL \url{http://arxiv.org/abs/1807.04863}.
\newblock arXiv: 1807.04863.

\bibitem[Donahue et~al.(2017)Donahue, Krähenbühl, and
  Darrell]{donahue_adversarial_2017}
Jeff Donahue, Philipp Krähenbühl, and Trevor Darrell.
\newblock Adversarial {Feature} {Learning}.
\newblock \emph{arXiv:1605.09782 [cs, stat]}, April 2017.
\newblock URL \url{http://arxiv.org/abs/1605.09782}.
\newblock arXiv: 1605.09782.

\bibitem[Goodfellow et~al.(2014)Goodfellow, Pouget-Abadie, Mirza, Xu,
  Warde-Farley, Ozair, Courville, and Bengio]{goodfellow_generative_2014}
Ian~J. Goodfellow, Jean Pouget-Abadie, Mehdi Mirza, Bing Xu, David
  Warde-Farley, Sherjil Ozair, Aaron Courville, and Yoshua Bengio.
\newblock Generative {Adversarial} {Networks}.
\newblock \emph{arXiv:1406.2661 [cs, stat]}, June 2014.
\newblock URL \url{http://arxiv.org/abs/1406.2661}.
\newblock arXiv: 1406.2661.

\bibitem[Gretton et~al.(2012)Gretton, Borgwardt, Rasch, Schölkopf, and
  Smola]{gretton_kernel_2012}
Arthur Gretton, Karsten~M. Borgwardt, Malte~J. Rasch, Bernhard Schölkopf, and
  Alexander Smola.
\newblock A {Kernel} {Two}-{Sample} {Test}.
\newblock \emph{Journal of Machine Learning Research}, 13\penalty0
  (25):\penalty0 723--773, 2012.
\newblock URL \url{http://jmlr.org/papers/v13/gretton12a.html}.

\bibitem[Grewal(2019)]{grewal_recent_2019}
Karan Grewal.
\newblock Recent trends and mutual information-based objectives in unsupervised
  learning, 2019.
\newblock URL \url{http://karangrewal.ca/blog/mutual-information-objectives/}.

\bibitem[Gulrajani et~al.(2016)Gulrajani, Kumar, Ahmed, Taiga, Visin, Vazquez,
  and Courville]{gulrajani_pixelvae_2016}
Ishaan Gulrajani, Kundan Kumar, Faruk Ahmed, Adrien~Ali Taiga, Francesco Visin,
  David Vazquez, and Aaron Courville.
\newblock {PixelVAE}: {A} {Latent} {Variable} {Model} for {Natural} {Images}.
\newblock \emph{arXiv:1611.05013 [cs]}, November 2016.
\newblock URL \url{http://arxiv.org/abs/1611.05013}.
\newblock arXiv: 1611.05013.

\bibitem[He et~al.(2019)He, Spokoyny, Neubig, and
  Berg-Kirkpatrick]{he_lagging_2019}
Junxian He, Daniel Spokoyny, Graham Neubig, and Taylor Berg-Kirkpatrick.
\newblock Lagging {Inference} {Networks} and {Posterior} {Collapse} in
  {Variational} {Autoencoders}.
\newblock \emph{arXiv:1901.05534 [cs, stat]}, January 2019.
\newblock URL \url{http://arxiv.org/abs/1901.05534}.
\newblock arXiv: 1901.05534.

\bibitem[He et~al.(2015)He, Zhang, Ren, and Sun]{he_deep_2015}
Kaiming He, Xiangyu Zhang, Shaoqing Ren, and Jian Sun.
\newblock Deep {Residual} {Learning} for {Image} {Recognition}.
\newblock \emph{arXiv:1512.03385 [cs]}, December 2015.
\newblock URL \url{http://arxiv.org/abs/1512.03385}.
\newblock arXiv: 1512.03385.

\bibitem[He et~al.(2020)He, Fan, Wu, Xie, and Girshick]{he_momentum_2020}
Kaiming He, Haoqi Fan, Yuxin Wu, Saining Xie, and Ross Girshick.
\newblock Momentum {Contrast} for {Unsupervised} {Visual} {Representation}
  {Learning}.
\newblock \emph{arXiv:1911.05722 [cs]}, March 2020.
\newblock URL \url{http://arxiv.org/abs/1911.05722}.
\newblock arXiv: 1911.05722.

\bibitem[Higgins et~al.(2016)Higgins, Matthey, Pal, Burgess, Glorot, Botvinick,
  Mohamed, and Lerchner]{higgins_beta-vae_2016}
Irina Higgins, Loic Matthey, Arka Pal, Christopher Burgess, Xavier Glorot,
  Matthew Botvinick, Shakir Mohamed, and Alexander Lerchner.
\newblock beta-{VAE}: {Learning} {Basic} {Visual} {Concepts} with a
  {Constrained} {Variational} {Framework}.
\newblock November 2016.
\newblock URL \url{https://openreview.net/forum?id=Sy2fzU9gl}.

\bibitem[Hoffman and Johnson(2016)]{hoffman_elbo_2016}
Matthew~D Hoffman and Matthew~J Johnson.
\newblock {ELBO} surgery: yet another way to carve up the variational evidence
  lower bound.
\newblock In \emph{{NeurIPS}}, page~4, 2016.

\bibitem[Kim et~al.(2018)Kim, Wiseman, Miller, Sontag, and
  Rush]{kim_semi-amortized_2018}
Yoon Kim, Sam Wiseman, Andrew~C. Miller, David Sontag, and Alexander~M. Rush.
\newblock Semi-{Amortized} {Variational} {Autoencoders}.
\newblock \emph{arXiv:1802.02550 [cs, stat]}, July 2018.
\newblock URL \url{http://arxiv.org/abs/1802.02550}.
\newblock arXiv: 1802.02550.

\bibitem[Kingma and Welling(2014)]{kingma_auto-encoding_2014}
Diederik~P. Kingma and Max Welling.
\newblock Auto-{Encoding} {Variational} {Bayes}.
\newblock \emph{arXiv:1312.6114 [cs, stat]}, May 2014.
\newblock URL \url{http://arxiv.org/abs/1312.6114}.
\newblock arXiv: 1312.6114.

\bibitem[Kingma and Welling(2019)]{kingma_introduction_2019}
Diederik~P. Kingma and Max Welling.
\newblock An {Introduction} to {Variational} {Autoencoders}.
\newblock \emph{Foundations and Trends® in Machine Learning}, 12\penalty0
  (4):\penalty0 307--392, 2019.
\newblock ISSN 1935-8237, 1935-8245.
\newblock \doi{10.1561/2200000056}.
\newblock URL \url{http://arxiv.org/abs/1906.02691}.
\newblock arXiv: 1906.02691.

\bibitem[Lake et~al.(2015)Lake, Salakhutdinov, and
  Tenenbaum]{lake_human-level_2015}
B.~Lake, R.~Salakhutdinov, and J.~Tenenbaum.
\newblock Human-level concept learning through probabilistic program induction.
\newblock \emph{Science}, 2015.
\newblock \doi{10.1126/science.aab3050}.

\bibitem[Larsen et~al.(2016)Larsen, Sønderby, Larochelle, and
  Winther]{larsen_autoencoding_2016}
Anders Boesen~Lindbo Larsen, Søren~Kaae Sønderby, Hugo Larochelle, and Ole
  Winther.
\newblock Autoencoding beyond pixels using a learned similarity metric.
\newblock \emph{arXiv:1512.09300 [cs, stat]}, February 2016.
\newblock URL \url{http://arxiv.org/abs/1512.09300}.
\newblock arXiv: 1512.09300.

\bibitem[Liu and Wang(2019)]{liu_stein_2019}
Qiang Liu and Dilin Wang.
\newblock Stein {Variational} {Gradient} {Descent}: {A} {General} {Purpose}
  {Bayesian} {Inference} {Algorithm}.
\newblock \emph{arXiv:1608.04471 [cs, stat]}, September 2019.
\newblock URL \url{http://arxiv.org/abs/1608.04471}.
\newblock arXiv: 1608.04471.

\bibitem[Makhzani et~al.(2016)Makhzani, Shlens, Jaitly, Goodfellow, and
  Frey]{makhzani_adversarial_2016}
Alireza Makhzani, Jonathon Shlens, Navdeep Jaitly, Ian Goodfellow, and Brendan
  Frey.
\newblock Adversarial {Autoencoders}.
\newblock \emph{arXiv:1511.05644 [cs]}, May 2016.
\newblock URL \url{http://arxiv.org/abs/1511.05644}.
\newblock arXiv: 1511.05644.

\bibitem[Mescheder et~al.(2017)Mescheder, Nowozin, and
  Geiger]{mescheder_adversarial_2017}
Lars Mescheder, Sebastian Nowozin, and Andreas Geiger.
\newblock Adversarial {Variational} {Bayes}: {Unifying} {Variational}
  {Autoencoders} and {Generative} {Adversarial} {Networks}.
\newblock January 2017.
\newblock URL \url{https://arxiv.org/abs/1701.04722v4}.

\bibitem[Oord et~al.(2016)Oord, Kalchbrenner, Vinyals, Espeholt, Graves, and
  Kavukcuoglu]{oord_conditional_2016}
Aaron van~den Oord, Nal Kalchbrenner, Oriol Vinyals, Lasse Espeholt, Alex
  Graves, and Koray Kavukcuoglu.
\newblock Conditional {Image} {Generation} with {PixelCNN} {Decoders}.
\newblock \emph{arXiv:1606.05328 [cs]}, June 2016.
\newblock URL \url{http://arxiv.org/abs/1606.05328}.
\newblock arXiv: 1606.05328.

\bibitem[Oord et~al.(2019)Oord, Li, and Vinyals]{oord_representation_2019}
Aaron van~den Oord, Yazhe Li, and Oriol Vinyals.
\newblock Representation {Learning} with {Contrastive} {Predictive} {Coding}.
\newblock \emph{arXiv:1807.03748 [cs, stat]}, January 2019.
\newblock URL \url{http://arxiv.org/abs/1807.03748}.
\newblock arXiv: 1807.03748.

\bibitem[Phuong et~al.(2018)Phuong, Welling, Kushman, Tomioka, and
  Nowozin]{phuong_mutual_2018}
Mary Phuong, Max Welling, Nate Kushman, Ryota Tomioka, and Sebastian Nowozin.
\newblock The {Mutual} {Autoencoder}: {Controlling} {Information} in {Latent}
  {Code} {Representations}.
\newblock February 2018.
\newblock URL \url{https://openreview.net/forum?id=HkbmWqxCZ}.

\bibitem[Poole et~al.(2019)Poole, Ozair, Oord, Alemi, and
  Tucker]{poole_variational_2019}
Ben Poole, Sherjil Ozair, Aaron van~den Oord, Alexander~A. Alemi, and George
  Tucker.
\newblock On {Variational} {Bounds} of {Mutual} {Information}.
\newblock \emph{arXiv:1905.06922 [cs, stat]}, May 2019.
\newblock URL \url{http://arxiv.org/abs/1905.06922}.
\newblock arXiv: 1905.06922.

\bibitem[Qian and Cheung(2019)]{qian_enhancing_2019}
Dong Qian and William~K. Cheung.
\newblock Enhancing {Variational} {Autoencoders} with {Mutual} {Information}
  {Neural} {Estimation} for {Text} {Generation}.
\newblock In \emph{Proceedings of the 2019 {Conference} on {Empirical}
  {Methods} in {Natural} {Language} {Processing} and the 9th {International}
  {Joint} {Conference} on {Natural} {Language} {Processing}
  ({EMNLP}-{IJCNLP})}, pages 4047--4057, Hong Kong, China, November 2019.
  Association for Computational Linguistics.
\newblock \doi{10.18653/v1/D19-1416}.
\newblock URL \url{https://aclanthology.org/D19-1416}.

\bibitem[Rezaabad and Vishwanath(2020)]{rezaabad_learning_2020}
Ali~Lotfi Rezaabad and Sriram Vishwanath.
\newblock Learning {Representations} by {Maximizing} {Mutual} {Information} in
  {Variational} {Autoencoders}.
\newblock \emph{arXiv:1912.13361 [cs, stat]}, January 2020.
\newblock URL \url{http://arxiv.org/abs/1912.13361}.
\newblock arXiv: 1912.13361.

\bibitem[Song and Ermon(2020)]{song_multi-label_2020}
Jiaming Song and Stefano Ermon.
\newblock Multi-label {Contrastive} {Predictive} {Coding}.
\newblock In \emph{{NeurIPS}}, July 2020.
\newblock URL \url{https://arxiv.org/abs/2007.09852v2}.

\bibitem[Srivastava et~al.(2017)Srivastava, Valkov, Russell, Gutmann, and
  Sutton]{srivastava_veegan_2017}
Akash Srivastava, Lazar Valkov, Chris Russell, Michael~U. Gutmann, and Charles
  Sutton.
\newblock {VEEGAN}: {Reducing} {Mode} {Collapse} in {GANs} using {Implicit}
  {Variational} {Learning}.
\newblock \emph{arXiv:1705.07761 [stat]}, November 2017.
\newblock URL \url{http://arxiv.org/abs/1705.07761}.
\newblock arXiv: 1705.07761.

\bibitem[Sugiyama et~al.(2012)Sugiyama, Suzuki, and
  Kanamori]{sugiyama_density-ratio_2012}
Masashi Sugiyama, Taiji Suzuki, and Takafumi Kanamori.
\newblock Density-ratio matching under the {Bregman} divergence: a unified
  framework of density-ratio estimation.
\newblock \emph{Annals of the Institute of Statistical Mathematics},
  64\penalty0 (5):\penalty0 1009--1044, October 2012.
\newblock ISSN 0020-3157, 1572-9052.
\newblock \doi{10.1007/s10463-011-0343-8}.
\newblock URL \url{http://link.springer.com/10.1007/s10463-011-0343-8}.

\bibitem[Suzuki et~al.(2008)Suzuki, Sugiyama, Sese, and
  Kanamori]{suzuki_approximating_2008}
Taiji Suzuki, Masashi Sugiyama, Jun Sese, and Takafumi Kanamori.
\newblock Approximating {Mutual} {Information} by {Maximum} {Likelihood}
  {Density} {Ratio} {Estimation}.
\newblock In \emph{New {Challenges} for {Feature} {Selection} in {Data}
  {Mining} and {Knowledge} {Discovery}}, pages 5--20. PMLR, September 2008.
\newblock URL \url{http://proceedings.mlr.press/v4/suzuki08a.html}.
\newblock ISSN: 1938-7228.

\bibitem[Tolstikhin et~al.(2019)Tolstikhin, Bousquet, Gelly, and
  Schoelkopf]{tolstikhin_wasserstein_2019}
Ilya Tolstikhin, Olivier Bousquet, Sylvain Gelly, and Bernhard Schoelkopf.
\newblock Wasserstein {Auto}-{Encoders}.
\newblock \emph{arXiv:1711.01558 [cs, stat]}, December 2019.
\newblock URL \url{http://arxiv.org/abs/1711.01558}.
\newblock arXiv: 1711.01558.

\bibitem[Tomczak and Welling(2017)]{tomczak_vae_2017}
Jakub~M. Tomczak and Max Welling.
\newblock {VAE} with a {VampPrior}.
\newblock In \emph{{AISTATS}}, May 2017.
\newblock URL \url{https://arxiv.org/abs/1705.07120v5}.

\bibitem[Tran et~al.(2017)Tran, Ranganath, and Blei]{tran_hierarchical_2017}
Dustin Tran, Rajesh Ranganath, and David~M. Blei.
\newblock Hierarchical {Implicit} {Models} and {Likelihood}-{Free}
  {Variational} {Inference}.
\newblock \emph{arXiv:1702.08896 [cs, stat]}, November 2017.
\newblock URL \url{http://arxiv.org/abs/1702.08896}.
\newblock arXiv: 1702.08896.

\bibitem[Tschannen et~al.(2020)Tschannen, Djolonga, Rubenstein, Gelly, and
  Lucic]{tschannen_mutual_2020}
Michael Tschannen, Josip Djolonga, Paul~K. Rubenstein, Sylvain Gelly, and Mario
  Lucic.
\newblock On {Mutual} {Information} {Maximization} for {Representation}
  {Learning}.
\newblock \emph{arXiv:1907.13625 [cs, stat]}, January 2020.
\newblock URL \url{http://arxiv.org/abs/1907.13625}.
\newblock arXiv: 1907.13625.

\bibitem[Uehara et~al.(2016)Uehara, Sato, Suzuki, Nakayama, and
  Matsuo]{uehara_generative_2016}
Masatoshi Uehara, Issei Sato, Masahiro Suzuki, Kotaro Nakayama, and Yutaka
  Matsuo.
\newblock Generative {Adversarial} {Nets} from a {Density} {Ratio} {Estimation}
  {Perspective}.
\newblock \emph{arXiv:1610.02920 [stat]}, November 2016.
\newblock URL \url{http://arxiv.org/abs/1610.02920}.
\newblock arXiv: 1610.02920.

\bibitem[Wang et~al.(2021)Wang, Blei, and Cunningham]{wang_posterior_2021}
Yixin Wang, David Blei, and John~P Cunningham.
\newblock Posterior {Collapse} and {Latent} {Variable} {Non}-identifiability.
\newblock In \emph{Advances in {Neural} {Information} {Processing} {Systems}},
  volume~34, pages 5443--5455. Curran Associates, Inc., 2021.
\newblock URL
  \url{https://proceedings.neurips.cc/paper/2021/hash/2b6921f2c64dee16ba21ebf17f3c2c92-Abstract.html}.

\bibitem[Wu et~al.(2018)Wu, Xiong, Yu, and Lin]{wu_unsupervised_2018}
Zhirong Wu, Yuanjun Xiong, Stella Yu, and Dahua Lin.
\newblock Unsupervised {Feature} {Learning} via {Non}-{Parametric}
  {Instance}-level {Discrimination}.
\newblock \emph{arXiv:1805.01978 [cs]}, May 2018.
\newblock URL \url{http://arxiv.org/abs/1805.01978}.
\newblock arXiv: 1805.01978.

\bibitem[Yang et~al.(2017)Yang, Hu, Salakhutdinov, and
  Berg-Kirkpatrick]{yang_improved_2017}
Zichao Yang, Zhiting Hu, Ruslan Salakhutdinov, and Taylor Berg-Kirkpatrick.
\newblock Improved {Variational} {Autoencoders} for {Text} {Modeling} using
  {Dilated} {Convolutions}.
\newblock \emph{arXiv:1702.08139 [cs]}, June 2017.
\newblock URL \url{http://arxiv.org/abs/1702.08139}.
\newblock arXiv: 1702.08139.

\bibitem[Zhao et~al.(2018)Zhao, Song, and Ermon]{zhao_infovae_2018}
Shengjia Zhao, Jiaming Song, and Stefano Ermon.
\newblock {InfoVAE}: {Information} {Maximizing} {Variational} {Autoencoders}.
\newblock \emph{arXiv:1706.02262 [cs, stat]}, May 2018.
\newblock URL \url{http://arxiv.org/abs/1706.02262}.
\newblock arXiv: 1706.02262.

\bibitem[Zhu et~al.(2020)Zhu, Huang, Li, Zhang, and Sun]{zhu_eqco_2020}
Benjin Zhu, Junqiang Huang, Zeming Li, Xiangyu Zhang, and Jian Sun.
\newblock {EqCo}: {Equivalent} {Rules} for {Self}-supervised {Contrastive}
  {Learning}.
\newblock \emph{arXiv:2010.01929 [cs]}, November 2020.
\newblock URL \url{http://arxiv.org/abs/2010.01929}.
\newblock arXiv: 2010.01929.

\end{thebibliography}

\clearpage

\appendix
\onecolumn


\begin{appendices}
    
    \section{Mutual Information}\label{appendix:mi}
    Recall that the mutual information across the variational joint is defined 
    
    \begin{equation}
        \begin{aligned}
        I_q(x ; z) &=D_{\mathrm{KL}}(q_\phi(x, z)|| p_\mathcal{D}(x) q(z)) \\
        &=\mathbb{E}_{q_\theta(x, z)}\left[\log \frac{q_\phi(x, z)}{p_\mathcal{D}(x) q_\phi(z)}\right]
        \end{aligned}        
    \end{equation}
    
    
    
    If the observations and the latents are independent, the mutual information is zero; in our case, we want to encourage the model to preserve their dependence, so we want it to be higher. Increasing $I_q(x;z)$ works against posterior collapse by preventing the posterior from always matching the prior (since this would not preserve any information). 
    We will show that the approach provides a new way to increase both of these MI measures that could be combined with existing approaches. 

    \section{Binary classifier and MI}\label{appendix:binclas}
    We use a categorical likelihood for multi-way classification to implement the critic. Another option that provides some intuition would be a binary classifier, that simply takes a pair and decides if they correspond or not in isolation. (See Appendix \ref{appendix:multirat} for the details of why the multisample case is preferred.) The connection between the binary classifier and MI follows directly from application of the density ratio trick \citep{sugiyama_density-ratio_2012}, which tells us a binary probabilistic classifier between two distributions estimates the density ratio between them. In our case, then, the optimal classifier would correspond to $\frac{ p(\textbf{z}, \textbf{x})}{p(\textbf{z})p(\textbf{x})}$. We highlight that in a different context, this same binary-classification density-ratio trick is what is used to power GANs: the discriminator estimates a density ratio between real and fake samples. In GANs, we do not want to be able to distinguish these distributions so we train the critic adversarially; here, we \textit{want} the critic to succeed. GANs also provide us some basis that we do not need to train the critic to optimality at every step, which would be too expensive - joint training of the critic and the model can yield the desired results \citep{goodfellow_generative_2014}. See Related Work for more discussion of GAN-related techniques.
    
    Thus, applying the density trick to our distributions at hand would provide us the integrand (of the MI expectation), and we could compute the expectation via Monte Carlo using all of the samples in the batch to get an estimate of the mutual information. The general technique of using a density ratio to estimate mutual information is introduced in \cite{suzuki_approximating_2008}, elaborated on in \cite{sugiyama_density-ratio_2012} and draws its roots to 2-sample testing via classifiers; we encourage the interested reader to refer to these for the history of the method. 
    
    \section{Multisample Density Ratio}\label{appendix:multirat}
    
    One practical reason we would be interested in using the information from all the samples is that the `one-sample' estimate of every density ratio term in the Monte Carlo expectation for MI described in \ref{appendix:binclas} will have very high variance; using information from multiple samples for each term and getting a `multi-sample' estimate would be more stable per \cite{poole_variational_2019}. When we use the multiclass objective pushing down the objective pushes up the MI implicitly, see Appendix \ref{appendix:multimi}. The tightness of this bound increases with the number of samples, so this is another reason we opt for the multisample approach. 
    
    
    \section{Mutual Information Bound}\label{appendix:multimi}
    
    This follows from analogy to CPC \citep{oord_representation_2019} (Appendix). This is an immediate application of the InfoNCE bound introduced there, which we follow here (along with \citep{grewal_recent_2019}); this is further elaborated on theoretically in \cite{poole_variational_2019}. 
    
    
    Consider a classifier that, for a latent sample $z_i$, tries to pick which observation $x$ from a set $X = \{x_1, \ldots, x_i, \ldots, x_K\}$ it corresponds to. (We can phrase the problem as the reverse as well - it doesn't matter since the density ratio and mutual information are symmetric.) We'll also follow the notation for the model critic (Equation \ref{eqn:sampling} left) for simplicity, but the inference critic (Equation \ref{eqn:sampling} right) follows the same steps. (Note we also drop subscripts on densities for clarity.)
    
    Consider Equation \ref{eqn:classification}. We know \citep{sugiyama_density-ratio_2012}, reshown by \citep{oord_representation_2019}, \citep{song_multi-label_2020}) that the classifier will estimate the density ratio up to a constant. That is, 
    \begin{equation}\label{eqn:densityprop}
        f(x, z) \propto \frac{p(x,z)}{p(x)p(z)} = \frac{p(x|z)}{p(x)}
    \end{equation}
    (where the second equality is a simple application of Bayes' rule.)
    
    We'll split the sum in the denominator of Equation \ref{eqn:classification} into 1) the term for the observation that corresponds to the latent at hand and 2) all the others. (In contrastive learning terminology, these are the positive and negatives respectively.) 
    
    \begin{equation}
    \begin{aligned}
        \mathcal{L} &= \mathbb{E}\left[\log \frac{f \left(x^{+}, z^{+}\right)}{\sum_{x \in S} f\left(x, z^{+}\right)}\right] \\
                    &= \mathbb{E}\left[\log \frac{f\left(x^{+}, z^{+}\right)}{f\left(x^{+}, z^{+}\right)+\sum_{x_j \in X \setminus x_i } f\left(x_{j}, z^{+}\right)}\right] \\
    \end{aligned}
    \end{equation}
    
    Since the classifier aims to estimate the density ratio (up to a constant), from Equation \ref{eqn:densityprop}
    
    \begin{equation}
    \begin{aligned}
                    &\approx \mathbb{E} \log \left[\frac{\frac{p\left(x^{+} \mid z^{+}\right)}{p\left(x^{+}\right)} C}{\frac{p\left(x^{+} \mid z^{+}\right)}{p\left(x^{+}\right)} C +\sum_{x_j \in X \setminus x_i} \frac{p\left(x_{j} \mid z^{+}\right)}{p\left(x_{j}\right)} C}\right] \\
                    &= \mathbb{E} \log \left[\frac{\frac{p\left(x^{+} \mid z^{+}\right)}{p\left(x^{+}\right)} }{\frac{p\left(x^{+} \mid z^{+}\right)}{p\left(x^{+}\right)}  +\sum_{x_j \in X \setminus x_i} \frac{p\left(x_{j} \mid z^{+}\right)}{p\left(x_{j}\right)} }\right]
    \end{aligned}
    \end{equation}
    
    We notice the `positive' term appears in the numerator and denominator. Doing some algebraic manipulation,

    \begin{equation}
        \begin{aligned}
                    &= \mathbb{E} \left( - \log \left[\frac{ \frac{p\left(x^{+} \mid z^{+}\right)}{p\left(x^{+}\right)}  +\sum_{x_j \in X \setminus x_i} \frac{p\left(x_{j} \mid z^{+}\right)}{p\left(x_{j}\right)} }{\frac{p\left(x^{+} \mid z^{+}\right)}{p\left(x^{+}\right)} }\right] \right) \\
                    &=\mathbb{E}\left(-\log \left[1+\frac{\sum_{x_j \in X \setminus x_i} \frac{p\left(x_{j} \mid z^{+}\right)}{p\left(x_{j}\right)}}{\frac{p\left(x^{+} \mid z^{+}\right)}{p\left(x^{+}\right)}}\right]\right) \\
                    &=\mathbb{E}\left(-\log \left[1+\frac{p\left(x^{+}\right)}{p\left(x^{+} \mid z^{+}\right)} \sum_{x_j \in X \setminus x_i} \frac{p\left(x_{j} \mid z^{+}\right)}{p\left(x_{j}\right)}\right]\right) \\
        \end{aligned}   
    \end{equation}
                    
    Now examine the sum over all terms but the `positive' one. This can be considered a (scaled) expectation of the density ratio over the `negative' terms - which should be $1$, as for independent $x$, $z$ the joint is the product of marginals. (Technically, since we are computing on samples, this is a Monte Carlo estimate of this expectation, but as noted by \cite{oord_representation_2019} it is nearly exact even with relatively low $K$; \cite{poole_variational_2019} shows a proof of the InfoNCE bound that does not use this approximation.)
    
    \begin{equation}
        \begin{aligned}
                    &\approx \mathbb{E}\left(-\log \left[1+\frac{p\left(x^{+}\right)}{p\left(x^{+} \mid z^{+}\right)} (K-1)\mathbb{E}_{x \sim X_{neg}} \left[ \frac{p\left(x \mid z^{+}\right)}{p\left(x\right)}\right] \right]\right) \\
                    &=\mathbb{E}\left(-\log \left[1+\frac{p\left(x^{+}\right)}{p\left(x^{+} \mid z^{+}\right)} (K-1)\right]\right) \\
                    &=\mathbb{E} \log \left[\frac{1}{1+\frac{p\left(x^{+}\right)}{p\left(x^{+} \mid z^{+}\right)} (K-1)}  \right] \\
                    & =\mathbb{E} \log \left[\frac{1}{K\frac{p\left(x^{+}\right)}{p\left(x^{+} \mid z^{+}\right)} - \frac{p\left(x^{+}\right)}{p\left(x^{+} \mid z^{+}\right)} + 1}  \right] \\
                    & =\mathbb{E} \log \left[\frac{1}{K\frac{p\left(x^{+}\right)}{p\left(x^{+} \mid z^{+}\right)} + \left(1 - \frac{p\left(x^{+}\right)}{p\left(x^{+} \mid z^{+}\right)}\right)} \right]
        \end{aligned}   
    \end{equation}
    
    
    
    \begin{equation}
        \begin{aligned}
                    & \leq \mathbb{E} \log \left[\frac{1}{K\frac{p\left(x^{+}\right)}{p\left(x^{+} \mid z^{+}\right)}} \right]\\
                    &= \mathbb{E} \log \left[\frac{1}{K} \frac{p\left(x^{+} \mid z^{+}\right)}{p\left(x^{+}\right)} \right]\\
                    &= \mathbb{E} \left( \log \left[\frac{p\left(x^{+} \mid z^{+}\right)}{p\left(x^{+}\right)} \right] \right) - \log K \\
                    &= I(x ; z)-\log K
        \end{aligned}   
    \end{equation}
    
    where the last line is clearly less than $I(x;z)$. Thus
    \begin{equation}
        I(x; z) \geq \mathcal{L} + \log K
    \end{equation}
    
    As the optimal loss (where the classifier is exactly a constant proportion of the density ratio) is bounded above by the MI, any suboptimal loss (which will be lower, since we are maximizing) will be bounded by the same. The key here is the $\log K$ term, which upper bounds the estimate of the MI (as $\mathcal{L} \leq 0$): intuitively, optimizing the loss pushes up the MI with a stick that is $\log K$ long. If the MI would otherwise fall to $0$, the regularization aims to increase it by up to $\log K$ - past this, the bound is loose (our stick cannot reach), so it is advantageous to use higher $K$. (This is why the binary case as a lower bound is not practical.) Empirically, we increase $I_q(x;z)$ by almost exactly $\log K$ with an inference-side critic, showing this technique works as well as the theory might tell us it can (Table \ref{table:textresults}, Table \ref{table:imgresults}).

    \section{Inference vs Model Critics} \label{appendix:critics}
    The model (`decoder'-side) critic corresponds to increasing $I_p(x;z)$ to prevent the likelihood from forgetting the latent, as previously discussed. For the inference (`encoder'-side) critic, the same analysis holds - instead of distinguishing the model joint from the product of the prior and the model approximate data distribution (marginal), it distinguishes the \textit{variational} joint $q_\phi (x, z)$ from the product of the empirical data distribution $p_{\mathcal{D}}(x)$ and the aggregate posterior $q(z)$ (whose samples are obtained by ancestral sampling, analogously to the samples from the model approximate data distribution). 
    Interestingly, adversarial variational Bayes \citep{mescheder_adversarial_2017} trains a similar critic adversarially, using this optimization to replace the ELBO. It also learns notably bad representations
    , so this is consistent, especially given that the ELBO terms they replace include the mutual information penalty (recall Equation \ref{eqn:elbo2}), but this could be interesting to consider in more depth.
    
    One disadvantage of the model critic is that it requires sampling from the model, which can be expensive for strong model networks like autoregressive ones - which are where it would have the most effect. The inference critic does not have this restriction. 
    
    \section{Experimental Protocol}\label{appendix:protocol}
    
    Protocol reproduced from \cite{he_lagging_2019}. 
    
    Text experiments: LSTM parameters are initialized from $\mathcal{U}(-0.01,0.01)$, with $\mathcal{U}(-0.1,0.1)$ for embedding parameters. The final hidden representation produced by the inference network is used to predict the latent variable with a linear transformation. The SGD optimizer is used with an initial learning rate of $1.0$, decayed by a factor of $2$ upon a validation loss plateau for at least 2 epochs. Training ends once the learning rate has been thus decayed 5 times. No text preprocessing is performed. Dropout of $0.5$ is used on the model network for the input embeddings and the pre-linear transformation output in vocabulary space.
    
    Image experiments: train/val/test splits are used identically to \cite{he_lagging_2019} and \cite{kim_semi-amortized_2018}. The Adam optimizer is used with an initial learning rate of 0.001, decayed by a factor of 2 upon a validation loss plateau for at least 2 epochs. Training ends once the learning rate has been thus decayed 5 times. Images are dynamically binarized -- that is, the input pixel values are considered parameters to Bernoulli random variables. Validation and test are performed with fixed binarization. The model network uses binary likelihood. The ResNet and PixelCNN are as described in \cite{he_lagging_2019}.
    
    \section{Mutual Information Comparison -- All Training} \label{appendix:miexpanded}
    \begin{figure}
    \centering
    \includegraphics[width=0.45\linewidth]{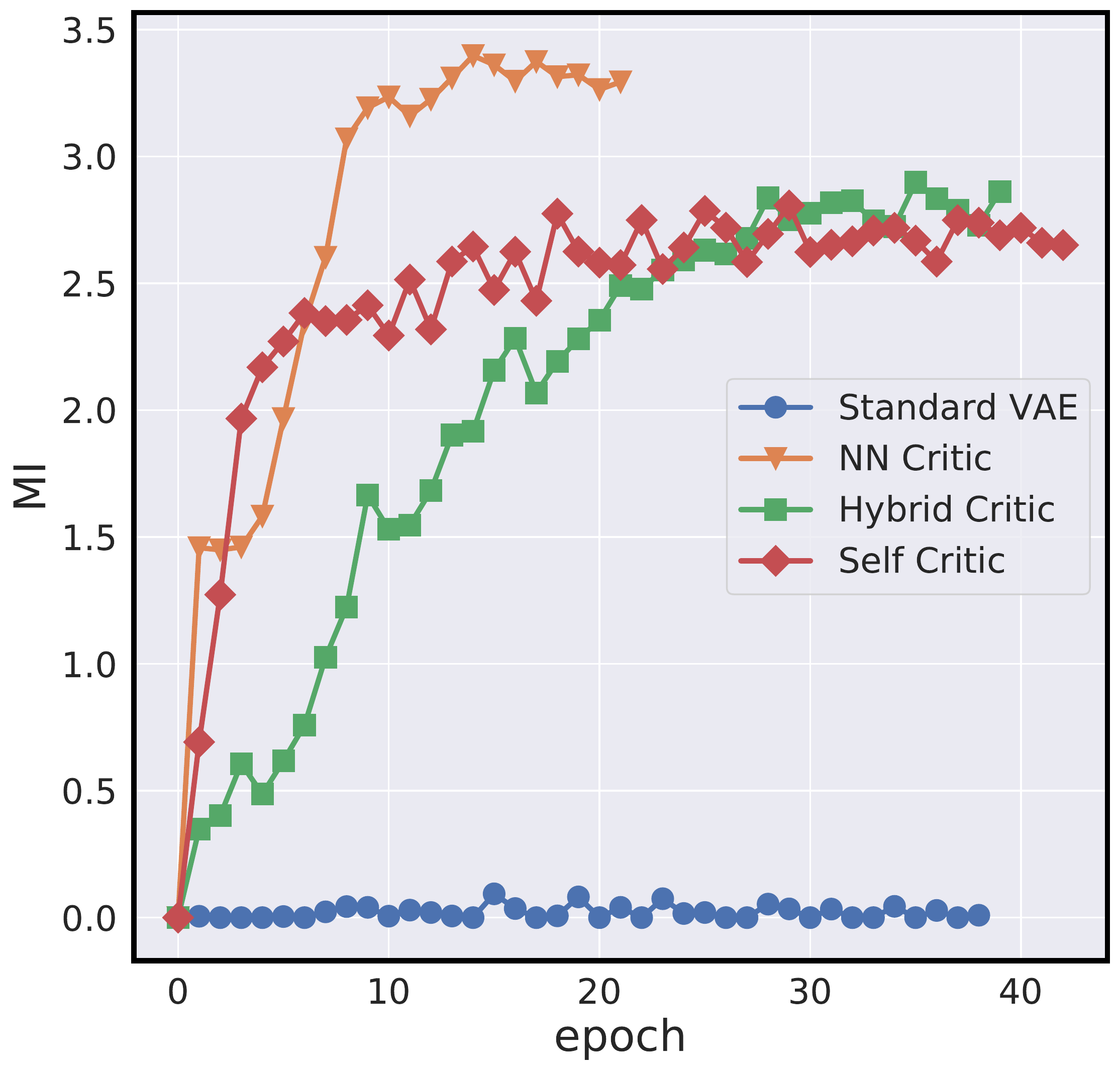}
    \caption{Comparison of mutual information across the variational family ($I_q$) for various critics vs baseline; the different endpoints are due to the termination condition for the experimental protocol depending on when a certain number of plateaus are reached.}
    \vspace{-2em}
    \label{fig:mi_comparison_expanded}
    \end{figure}

    \section{Discussion of VAE-MINE}\label{appendix:vaemine}
    Intuitively, our inference critics solve a classification task with a simple cross-entropy loss. This can be optimized with vanilla backprop. VAE-MINE adds a different term, based on MINE, to train an energy function that does not solve the same task; to optimize it, they resort to Taylor approximations and convex duality. This only implicitly results in contrasting the two distributions, while we directly train our inf. critic to do so. Yet, (per \cite{poole_variational_2019} Sec 2.2,) their way of estimating MINE \textit{is not even a correct bound on the MI}. Even if we ignore this, they lose the critical aspect of speed; for every size-$n$ batch, their bound uses $n^2$ forward passes (\citep{poole_variational_2019} Sec 3) vs our $2n$ (vs base VAE's $n$). This scales poorly. (There is no code available for VAE-MINE for empirical comparison, but there is a decisive gap between their quadratic and our linear runtime.) Finally, (\citep{poole_variational_2019} App. A), the bound we use is lower variance than (the correct) MINE. Our method is theoretically appealing, correct, and fast.
    
    \end{appendices}
    
\end{document}